\documentclass{article}

\usepackage[preprint]{corl_2026} 

\usepackage{amsmath}
\usepackage{amssymb}
\usepackage{booktabs}
\usepackage{graphicx}
\usepackage{multirow}
\usepackage{array}
\usepackage[table]{xcolor}
\usepackage{wrapfig}
\usepackage{enumitem}
\usepackage{algorithm}
\usepackage{algpseudocode}
\usepackage{xspace}
\usepackage{pgfplots}
\pgfplotsset{compat=1.18}
\usepackage{capt-of}

\definecolor{citecolor}{HTML}{0071BC}
\definecolor{gapblue}{HTML}{4682B4} 
\hypersetup{colorlinks,linkcolor={red},citecolor={citecolor}}

\title{TempoVLA: Learning Speed-Controllable\\ Vision-Language-Action Policies}

\author{
  Dong Jing$^{12*\ddagger}$, Jingchen Nie$^{2*\ddagger}$, Tianqi Zhang$^{2*\ddagger}$, Jiaqi Liu$^{2}$, \\
  \textbf{Huaxiu Yao$^{2}$, Zhiwu Lu$^{1\dagger}$, Mingyu Ding$^{2\dagger}$}\\
  $^{1}$RUC, $^{2}$UNC\\
  \texttt{\{jingdong98, luzhiwu\}@ruc.edu.cn, md@cs.unc.edu}\\
  \footnotesize{$^{*}$Equal contribution  $^{\dagger}$Corresponding authors}
}

\begin{document}
\maketitle
\begingroup
\renewcommand{\thefootnote}{\fnsymbol{footnote}}
\footnotetext[3]{This work was done during an internship at UNC.}
\endgroup

\vspace{-0.15in}
\begin{figure}[h]
  \centering
  \includegraphics[width=0.99\linewidth]{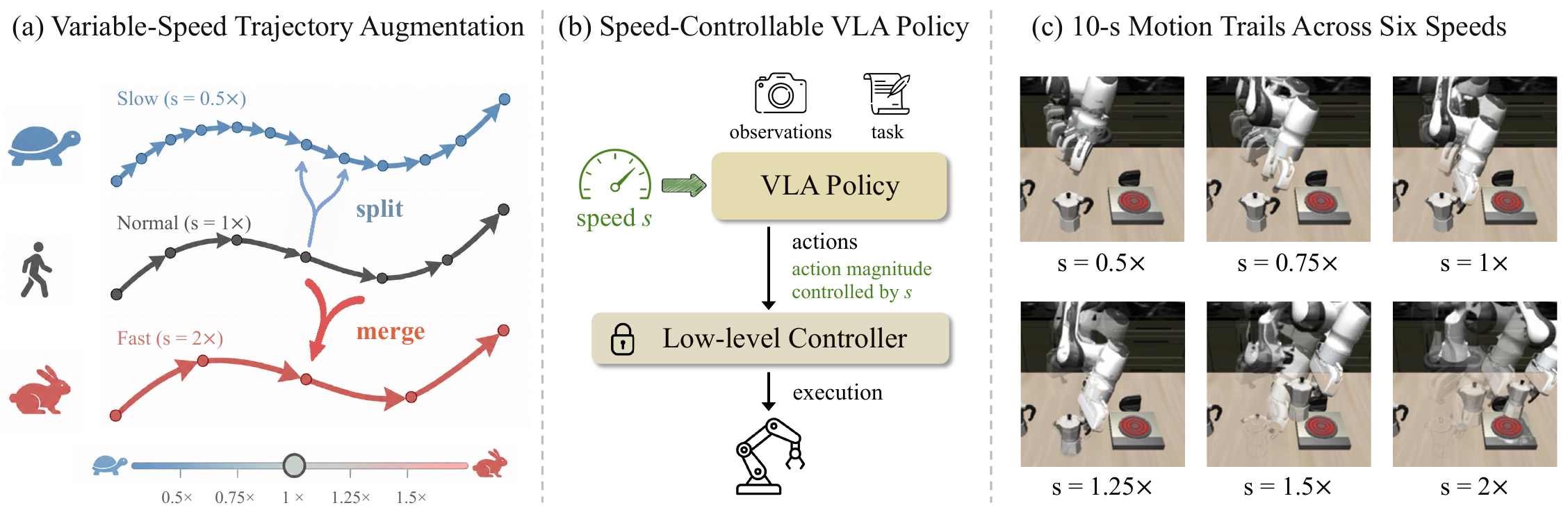}
  \caption{\textbf{TempoVLA: a speed-controllable VLA framework.}
  \textit{(a)}~VSTA re-times any demonstration to a target speed by selectively \emph{merging} consecutive actions to speed up or \emph{splitting} them to slow down, while preserving motion semantics.
  \textit{(b)}~The policy takes a scalar speed $s$ as an explicit conditioning input that scales the magnitude of its predicted actions, with the low-level controller left unchanged.
  \textit{(c)}~For a fixed task, the rollout motion trails of one TempoVLA policy at six commanded speeds tighten under slow commands and stretch under fast ones.}
  \label{fig:teaser}
  \vspace{-0.1in}
\end{figure}

\begin{abstract}
Robot manipulation alternates between low-risk transit phases that call for fast execution and high-risk contact stages that demand slow, precise motion. 
Yet existing Vision-Language-Action models (VLAs) only inherit a single fixed speed from training demonstrations.
Prior efforts to accelerate VLAs through model compression, KV-cache reuse, or reinforcement learning only shift the policy from one fixed speed to another, and leave deceleration almost unexplored.
We observe that the magnitude of each predicted action already governs how fast the robot moves, opening a direct route to controllable execution speed. 
We turn this observation into TempoVLA, a single VLA whose execution speed is controlled by an explicit condition. 
TempoVLA combines two coupled components.
(1) A data-side Variable-Speed Trajectory Augmentation (VSTA) that re-times demonstration to any target speed by merging or splitting actions while preserving its motion semantics. 
(2) A model-side conditioning mechanism that feeds the speed to the policy.
Statistics show that VSTA reaches the requested speed with negligible motion error. 
Experiments in simulation and on real-world tasks demonstrate that TempoVLA achieves flexible speed control in both directions, while VSTA additionally boosts the default $1\times$ performance via better data utilization. 
Furthermore, by cooperating with a large multimodal model, TempoVLA realizes dynamic speed control, accelerating through low-risk phases and decelerating for high-risk ones.
\end{abstract}


\section{Introduction}
\label{sec:intro}
\vspace{-0.05in}

Vision-Language-Action models (VLAs) have emerged as a mainstream paradigm for general-purpose robot manipulation~\citep{rt2,openvla,pi0,pi05,octo,rdt}.
By training large vision-language backbones on robot demonstrations, VLAs follow language instructions and act across diverse embodied platforms, from robot arms to quadrupeds and humanoids.

A core but under-controlled dimension in deploying these policies is the execution speed.
Real manipulation alternates between low-risk transit phases that should run fast and high-risk contact phases that should slow down for precision.
Yet today's VLAs silently inherit a single fixed execution speed from their training demonstrations.
Existing efforts to alter this speed sit at the inference or controller side, accelerating policies through model compression, KV-cache reuse, asynchronous action chunking, or Reinforcement-Learning (RL) rollouts~\citep{tinyvla,smolvla,yang2025efficientvla,pertsch2025fast,kim2025openvlaoft,black2025rtc,liu2026legato,yuan2025speedtuning,ma2025realtimevla,yang2026realtimevlav2,song2026fastdvla,park2024pte,wu2026speeduppatch}.
However, these methods merely shift the policy from one fixed speed to another rather than offering explicit, on-demand speed control.
They also focus exclusively on acceleration, while deceleration, which remains essential for precision insertion, fragile handover, and other contact-rich behaviors, receives little attention.
The open challenge is therefore to give a single VLA explicit, bidirectional speed control without retraining its base architecture from scratch.

We observe that the magnitude of each predicted action already governs how fast the robot moves in the embodied setting, which opens a direct route to controllable execution speed.
As shown in Figure~\ref{fig:teaser}, building on this insight, we approach the problem from two coupled sides while leaving the low-level controller untouched.
On the data side, we introduce \emph{Variable-Speed Trajectory Augmentation} (VSTA), an online strategy that re-times any existing demonstration to any target speed by \emph{merging} consecutive actions into fewer, larger ones to speed up, or \emph{splitting} actions into more, smaller ones to slow down, while preserving the underlying motion semantics (Figure~\ref{fig:teaser}\,(a)).
On the model side, we feed the speed $s$ to the policy as an explicit conditioning input that scales its predicted action magnitude through three different injection schemes (Figure~\ref{fig:teaser}\,(b)).
Both our data-side and model-side designs are lightweight and applicable to all existing VLAs.

Figure~\ref{fig:teaser}\,(c) previews the resulting behavior: a single policy trained with our method can execute the task at multiple different commanded speeds, with the motion trail tightening at slow speeds and stretching at fast ones.
Experiments on LIBERO and on real-world tasks confirm that this control extends in both directions, and that VSTA additionally acts as useful data augmentation that improves default $1\times$ performance.

We further show that pairing the speed-conditioned policy with a Vision-Language Model (VLM) enables automated \textit{dynamic speed scheduling} and boosts better performance, where the system accelerates through low-risk phases and decelerates for high-risk ones without human intervention.

In summary, our contribution is threefold.
\begin{enumerate}[leftmargin=*,itemsep=2pt,topsep=2pt]
    \item We propose VSTA together with speed conditioning, a lightweight data-and-model pair that equips existing VLAs with bidirectional speed control without new data collection.
    \item We find that with properly re-timed data, speed control is easy to implant and largely independent of the conditioning mechanism, and that variable-speed training acts as an effective augmentation that consistently lifts the default 1x success rate in simulation and the real world.
    \item We demonstrate that this design extends to VLM-driven dynamic speed scheduling, turning execution speed into a new control channel for higher-level reasoners.
\end{enumerate}

\vspace{-0.1in}
\section{Related Work}
\label{sec:related}
\vspace{-0.1in}

\noindent\textbf{Vision-Language-Action Models.}
Vision-Language-Action models (VLAs) map visual observations and language instructions to executable robot actions~\citep{rt1,rt2,openvla,pi0,pi05,octo,rdt,palme,vima,smolvla,gato,baku,robocat,qu2025spatialvla,li2024cogact,deng2025graspvla,dreamvla25,cheang2025gr,bjorck2025gr00t,univla,zheng2025xvla,mu2024embodiedgpt,starvla2025,jing2025mixture_of_horizons}.
They are trained at scale by imitating large collections of teleoperated demonstrations, supported by a growing ecosystem of embodied datasets and manipulation benchmarks~\citep{openx,droid,bridgedatav2,libero,robomimic,metaworld,mimicgen,robonet,rlbench,robocasa365,bu2025agibot}.
Across this landscape, the action decoder broadly falls into three families: regression heads that emit continuous actions directly, as in ACT~\citep{act}; diffusion or flow-matching heads that model the action distribution generatively, as in Diffusion Policy~\citep{diffusionpolicy}, ALOHA Unleashed~\citep{alohaunleashed}, and $\pi_0$~\citep{pi0}; and discrete token heads that autoregressively decode action tokens, as in RT-2~\citep{rt2} and OpenVLA~\citep{openvla}.
Yet regardless of decoder family, the execution speed of a trained VLA is silently inherited from its demonstration data, which becomes a bottleneck when a single task contains phases that call for different motion paces.

\noindent\textbf{Model-based VLA Acceleration.}
A first line of work makes VLAs faster by intervening inside the policy itself.
\emph{Model compression} shrinks the policy footprint to reduce per-step inference cost, as in TinyVLA~\citep{tinyvla}, SmolVLA~\citep{smolvla}, and EfficientVLA~\citep{yang2025efficientvla}.
\emph{Token and KV-cache compression} accelerates the language backbone or the action head, such as the FAST action tokenizer for $\pi_0$~\citep{pertsch2025fast} and the parallel-decoding fine-tune of OpenVLA-OFT~\citep{kim2025openvlaoft}.
\emph{Asynchronous chunked execution} hides inference latency by overlapping prediction with execution, as in Real-Time Chunking~\citep{black2025rtc}, while related work trains flow policies to produce smoother chunk-boundary continuations that remove the discontinuities at chunk transitions~\citep{liu2026legato}.
\emph{Reinforcement-learning fine-tuning} retrains the policy with task rewards to encourage faster, more decisive behavior~\citep{yuan2025speedtuning}.
A complementary line manipulates the demonstration tempo itself, including DemoSpeedup~\citep{guo2025demospeedup}, SpeedAug~\citep{nam2025speedaug}, ESPADA~\citep{kim2025espada}, and SAIL~\citep{arachchige2025sail}.
However, none of these methods expose execution speed as an explicit, on-demand control; they at best shift the policy from one fixed speed to another.
Deceleration, in particular, is left almost entirely unaddressed.

\noindent\textbf{Model-free VLA Acceleration.}
A complementary line operates strictly downstream of the policy, tuning the robot-side execution stack for stable and rapid motion without touching policy weights.
Classical and GPU-accelerated motion planners such as CHOMP~\citep{ratliff2009chomp}, Riemannian motion policies~\citep{ratliff2018riemannian}, and cuRoBo~\citep{sundaralingam2023curobo} produce smoother, faster trajectory tracking.
Recent work further shows that low-level controller gains themselves substantially shape how a learned policy executes its predictions~\citep{bronars2026tune}.
These approaches are orthogonal to ours and can be stacked on top of a speed-conditioned VLA, but operating strictly downstream they can only rescale or smooth whatever actions the policy emits.
They cannot recover from upstream pathologies such as imprecise predictions or hesitation stalls inherited from teleoperation data.


\section{Methodology}
\label{sec:method}

\subsection{Problem Formulation}
\label{sec:method-formulation}

\noindent\textbf{VLA for robot manipulation.}
A Vision-Language-Action (VLA) policy $\pi_\theta$ is a sequential decision model for end-to-end robot manipulation.
At each step $t$, it consumes the observation $o_t = (v_t, \ell_t, \rho_t)$ comprising the visual input $v_t$, the language instruction $\ell_t$, and an optional proprioceptive state $\rho_t$.
From this observation, the policy predicts an action chunk $A_t = (a_t, \dots, a_{t+H-1})$ of horizon $H$.
The policy is trained by imitation on a demonstration set $\mathcal{D} = \{(o_t, A_t)\}$:
\begin{equation}
\small
\theta^\star = \arg\min_{\theta}\; \mathbb{E}_{(o_t, A_t)\sim \mathcal{D}}\big[\mathcal{L}\big(\pi_\theta(o_t),\, A_t\big)\big],
\label{eq:bc}
\end{equation}
where $\mathcal{L}$ is the imitation objective (e.g., regression or flow-matching).

\noindent\textbf{Goal of TempoVLA.}
TempoVLA aims to produce a single policy whose execution speed is controllable through an explicit scalar input.
We split this goal into two coupled sub-objectives.
On the data side, we want to online re-time any demonstration in $\mathcal{D}$ to an arbitrary target speed $s \in \mathbb{R}_{+}$ without losing motion semantics, yielding a multi-speed augmented dataset $\widetilde{\mathcal{D}}$.
On the model side, we want a speed-conditioned policy $\pi_\theta(o_t, s)$ trained on $\widetilde{\mathcal{D}}$ that scales its predicted action magnitudes according to $s$, where $s>1$ speeds up, $s<1$ slows down, and $s=1$ recovers the default speed.
The downstream low-level controller is left untouched throughout.

\begin{figure}[t]
  \centering
  \vspace{-12pt}
  \includegraphics[width=0.99\linewidth]{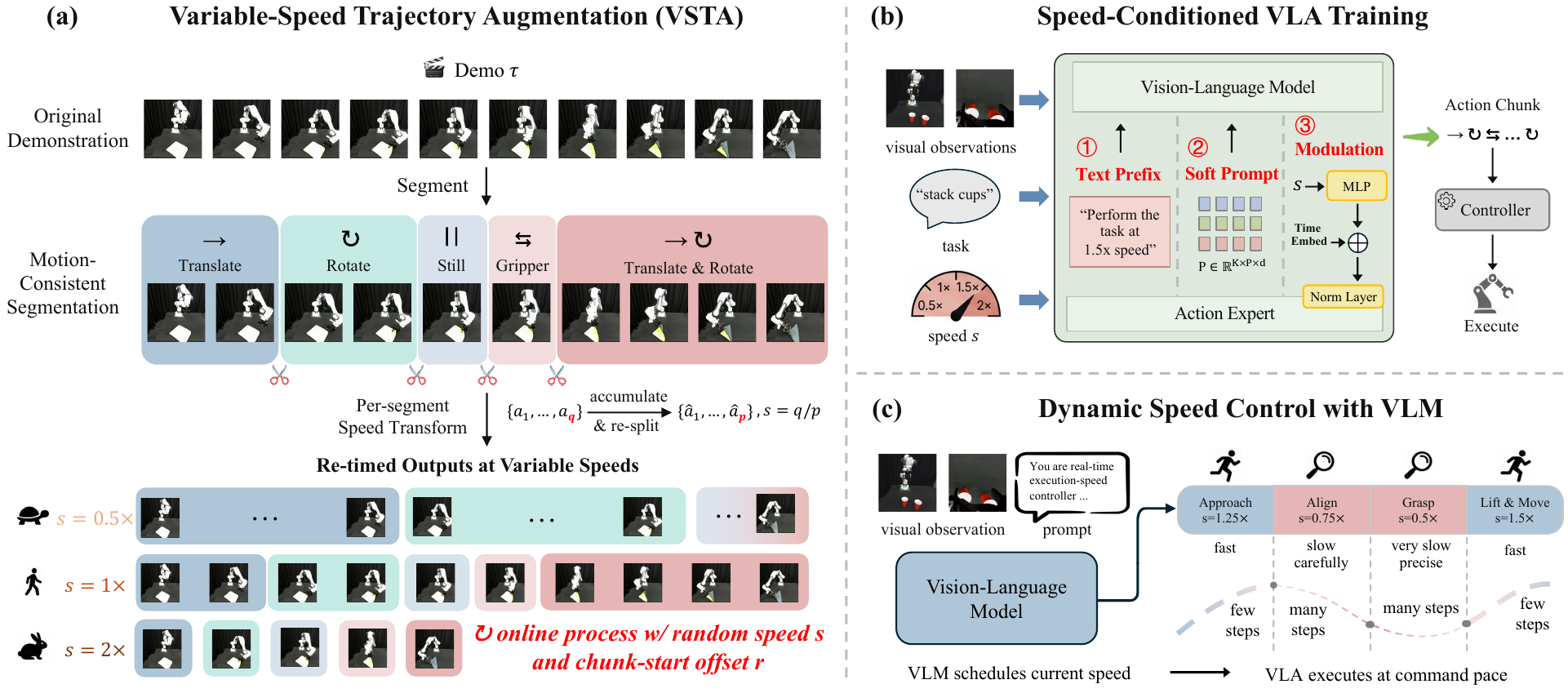}
  \vspace{-6pt}
  \caption{\textbf{Framework of TempoVLA.}
  \textit{(a)}~VSTA re-splits each motion-consistent segment from $q$ to $p$ actions to realize $s=q/p$, with $s$ and a chunk-start offset $r$ resampled online.
  \textit{(b)}~The speed $s$ enters the policy via a \emph{text prefix}, a \emph{soft prompt}, or an MLP-driven \emph{modulation}.
  \textit{(c)}~At deployment, a VLM scheduler observes the scene and dispatches per-chunk speeds for TempoVLA to execute.}
  \label{fig:framework}
  \vspace{-16pt}
\end{figure}

\subsection{Variable-Speed Trajectory Augmentation}
\label{sec:method-vsta}

Variable-Speed Trajectory Augmentation (VSTA) realizes the data-side objective of TempoVLA by re-timing any demonstration to a target speed $s$ online during training, as illustrated in Figure~\ref{fig:framework}\,(a).
The procedure has three steps and is detailed in Algorithm~\ref{alg:vsta} (Appendix~\ref{sec:appendix-vsta-alg}): motion-consistent \emph{segmentation}, chunk-level \emph{speed transform}, and \emph{online chunk-start sampling}.

\vspace{-2pt}
\noindent\textbf{Motion-consistent segmentation.}
We first cut each demonstration into segments whose motion is internally consistent.
Every frame is labeled with one of four motion modes (\emph{still}, \emph{translate}, \emph{rotate}, or \emph{translate-and-rotate}) according to whether its translation and rotation magnitudes exceed a small threshold, and a boundary is placed at every mode change.
Within a single mode, we further split whenever the motion direction reverses, i.e., the cosine similarity between consecutive translation or rotation directions falls below $\tau_{\mathrm{dir}}$.
Gripper open/close events are kept as hard boundaries so that a discrete state switch is never blurred by resampling.

\vspace{-2pt}
\noindent\textbf{Chunk-level speed transform.}
Inside each segment we realize the target speed $s$ by re-allocating actions between source and output frames.
We write $s = q/p$ with coprime integers $q, p$, so that $q$ source frames are mapped to $p$ output frames ($q > p$ speeds up, $q < p$ slows down).
We partition the segment into non-overlapping chunks of $q$ consecutive frames, leaving any trailing remainder shorter than $q$ unchanged.
For each chunk, we accumulate its total motion $\Delta = \sum_{i=1}^{q} a_i$, and then re-split $\Delta$ into $p$ equal-magnitude steps by linearly interpolating the cumulative motion.
By construction, the $p$ new actions sum back to $\Delta$ exactly, so the integrated motion of the chunk is preserved and only the within-chunk shape is altered.

\vspace{-2pt}
The accumulate-then-split operation is valid only when adding actions equals composing them.
This holds for Cartesian translation in $\mathbb{R}^3$, joint velocities, and rotational increments written as axis-angle vectors in $\mathfrak{so}(3)$ (whose axis the segmentation keeps approximately constant within a segment).
Representations that are not closed under addition, such as unit quaternions, rotation matrices, or Euler angles, must first be mapped to $\mathfrak{so}(3)$ or interpolated on the manifold (e.g., SLERP~\citep{shoemake1985animating}) before VSTA can be applied.
The gripper command is copied discretely, and gripper switches serve as anchors so they are never averaged across.

\vspace{-2pt}
\noindent\textbf{Online chunk-start sampling.}
Once a chunk is sped up, only the observation at the chunk start corresponds to an emitted action, and the other $q - 1$ observations would have to be dropped from training.
To avoid permanently discarding them, following~\cite{guo2025demospeedup}, we randomize where the chunks begin: for each segment we sample an offset $r \sim \mathcal{U}\{0, \dots, q-1\}$, so the first $r$ frames pass through verbatim and the chunks start at $r, r+q, r+2q, \dots$
Because VSTA runs online during training, a fresh offset is drawn every time the demonstration is sampled.
Over the course of training, every source frame eventually becomes a chunk start and contributes a valid training observation.
\vspace{-4pt}
\subsection{Speed Conditioning in TempoVLA}
\label{sec:method-conditioning}
\vspace{-4pt}

On top of the multi-speed dataset $\widetilde{\mathcal{D}}$ produced by VSTA, we train TempoVLA as a speed-conditioned policy $\pi_\theta(o_t, s)$ through one of three lightweight schemes that inject $s$ into a VLA (Figure~\ref{fig:framework}\,(b)).

\noindent\textbf{Textual prefix.}
We prepend a short phrase such as ``\texttt{Perform the task at $\langle s\rangle$x speed.}'' to the original instruction $\ell$, leaving the architecture entirely unchanged.

\noindent\textbf{Speed-modulated RMSNorm.}
A small two-layer MLP $\phi_{\mathrm{mod}}(s) \in \mathbb{R}^{d_{\mathrm{mod}}}$ embeds the scalar speed, and we add its output to the flow-matching timestep embedding $\sigma_{\mathrm{ts}}$ that already conditions every transformer block of the action expert.
The summed signal drives RMSNorm of each expert layer,
\begin{equation}
\small
\mathrm{adaRMSNorm}\bigl(x;\,\sigma_{\mathrm{ts}} + \phi_{\mathrm{mod}}(s)\bigr) \;=\; \gamma\bigl(\sigma_{\mathrm{ts}} + \phi_{\mathrm{mod}}(s)\bigr) \odot \frac{x}{\lVert x \rVert_{\mathrm{RMS}}},
\end{equation}
so that $s$ rescales the feature statistics throughout the expert.

\noindent\textbf{Soft prompt with speed anchors.}
We maintain a learnable tensor $\mathbf{P} \in \mathbb{R}^{K \times P \times d_{\mathrm{emb}}}$ that stores $P$ soft-prompt tokens for each of $K$ training-speed anchors $s_k \in \mathcal{S}$.
During training, the $P$ tokens for the current speed are inserted between the image and language tokens at the encoder input.
At inference, we pick the anchor nearest the requested speed, $k^\star = \arg\min_k \lvert s - s_k \rvert$, and use its tokens.

\subsection{Dynamic Speed Control with a VLM Scheduler}
\label{sec:method-dynamic}

Beyond fixed speed commands, TempoVLA supports automated \textit{dynamic speed scheduling} when paired with a high-level Vision-Language Model (VLM), as illustrated in Figure~\ref{fig:framework}\,(c).
At deployment, the VLM takes the current observations and a prompt as input, and predicts the speed $s_t$ for the next few action chunks.
TempoVLA then executes those chunks at the dispatched speed $s_t$.
The behavior accelerates through low-risk transit phases such as free-space approach and slows down for high-risk contact phases such as grasping or insertion.
Because the VLM and TempoVLA communicate only through the scalar $s$, the planner can be upgraded without retraining the policy.


\section{Simulation Experiments}
\label{sec:sim-exp}


\noindent\textbf{Simulation Setup.}
We evaluate TempoVLA on LIBERO~\citep{libero}, which provides four manipulation task suites (Spatial, Object, Goal, Long), each containing 10 tasks and 500 human-teleoperated demonstrations.
Its demonstrations are smooth and free of abrupt speed changes, which makes it a clean testbed for speed control.
Each action is a $7$-dim end-effector (EEF) command comprising a translation $(\Delta x, \Delta y, \Delta z)$, an axis-angle rotation increment, and a gripper signal.
The translation and rotation parts lie in the linearly composable space of Section~\ref{sec:method-vsta}, so VSTA applies through its accumulate-then-split operation, while the gripper is handled discretely.

\noindent\textbf{Base Model and Implementation Details.}
Our base model is $\pi_{0.5}$~\citep{pi05}, a flow-matching VLA built on PaliGemma~\citep{paligemma} and pre-trained on large-scale embodied datasets.
We feed the target speed $s$ with the textual prefix as the default unless stated otherwise.
All models are trained for 30k iterations with batch size 512 on 32 NVIDIA H20 GPUs under a fixed random seed for fair comparison.

\subsection{Feasibility of Variable-Speed Trajectory Augmentation}
\label{sec:sim-feasibility}

We first verify that VSTA produces executable demonstrations at each target speed.
For each $s \in \{0.5, 0.75, 1, 1.25, 1.5, 2\}$, we apply VSTA to the LIBERO demonstrations and replay the re-timed actions in the simulator.
The segmentation stage is speed-independent and divides each demonstration into $5.96$ segments of mean length $41$ steps on average.
The default $1\times$ replays the original actions and serves as the baseline, so its motion error is reported as ``--''.
As Table~\ref{tab:vsta-feasibility} shows, the realized \emph{Data Ratio} closely tracks the target speed, with a small rounding gap appearing at higher acceleration ratios because the per-segment chunk count must be integer-valued.
For replay success rate (\emph{SR}), speeds close to the baseline stay highly reliable, with $0.75\times$ and $1.25\times$ reaching $92.9\%$ and $92.4\%$ versus $97.6\%$ at $1\times$.
The SR then degrades monotonically as the target moves further from $1\times$ in either direction.
The \emph{Motion Err.}, the absolute deviation in integrated end-effector displacement caused by re-timing, grows with the speed factor $s$ but stays below $5\times 10^{-8}$ throughout, which is negligible compared to controller tolerances.
Overall, VSTA is a reliable data-processing primitive for producing variable-speed demonstrations to train TempoVLA.

\begin{table}[t]
\centering
\begin{minipage}[t]{0.5\textwidth}
\centering
\small
\setlength{\tabcolsep}{3.5pt}
\renewcommand{\arraystretch}{1.15}
\caption{\textbf{Feasibility of VSTA on LIBERO.} Re-timed demonstrations replay at each target speed $s$. Blue subscripts give the Data Ratio gap to $s$.}
\vspace{-6pt}
\label{tab:vsta-feasibility}
\begin{tabular}{c c c c c}
\toprule
Target $s$ & Data Ratio & Succ.\,(\#) & SR\,(\%) & Motion Err. \\
\midrule
0.5  & 0.5  & 664 & 83.0 & 2.8E-10 \\
0.75 & 0.76\textsubscript{\textcolor{gapblue}{$+0.01$}} & 743 & 92.9 & 4.4E-9 \\
\rowcolor{gray!15} \textbf{1} & \textbf{1} & \textbf{781} & \textbf{97.6} & \textbf{--} \\
1.25 & 1.20\textsubscript{\textcolor{gapblue}{$-0.05$}} & 739 & 92.4 & 1.1E-8 \\
1.5  & 1.43\textsubscript{\textcolor{gapblue}{$-0.07$}} & 653 & 81.6 & 2.2E-8 \\
2    & 1.90\textsubscript{\textcolor{gapblue}{$-0.10$}} & 540 & 67.5 & 4.8E-8 \\
\bottomrule
\end{tabular}
\end{minipage}
\hfill
\begin{minipage}[t]{0.47\textwidth}
\centering
\footnotesize
\setlength{\tabcolsep}{3pt}
\caption{\textbf{Ablation of speed-integration scheme.} \emph{SR}: average success rate (\%); \emph{Steps}: average rollout length on successes.}
\vspace{-6pt}
\label{tab:speed-integration}
\begin{tabular}{l cc cc cc}
\toprule
\multirow{2}{*}{Speed} & \multicolumn{2}{c}{Text} & \multicolumn{2}{c}{Modulation} & \multicolumn{2}{c}{Soft Prompt-8} \\
\cmidrule(lr){2-3} \cmidrule(lr){4-5} \cmidrule(lr){6-7}
 & SR\,$\uparrow$ & Steps & SR\,$\uparrow$ & Steps & SR\,$\uparrow$ & Steps \\
\midrule
$0.75\times$ & 96.5 & 197 & 95.8 & 198 & 96.4 & 200 \\
$1.0\times$  & 96.9 & 151 & 97.0 & 152 & 95.8 & 153 \\
$1.25\times$ & 97.0 & 126 & 97.3 & 129 & 96.8 & 129 \\
$1.5\times$  & 96.8 & 111 & 97.2 & 111 & 96.9 & 111 \\
\midrule
Avg. & 96.8 & 146.6 & 96.8 & 147.6 & 96.5 & 148.2 \\
\bottomrule
\end{tabular}
\end{minipage}
\vspace{-0.1in}
\end{table}

\subsection{Ablation on the Speed-Integration Scheme}
\label{sec:sim-integration}

We next study how the speed signal should be injected into the VLA.
We compare the three schemes of Section~\ref{sec:method-conditioning}: a textual prefix (\emph{Text}), an action-expert modulation (\emph{Modulation}), and a soft prompt with $P{=}8$ anchor tokens (\emph{Soft Prompt-8}).
All three are trained and evaluated on LIBERO with the same speed set $\{0.75, 1, 1.25, 1.5\}\times$, and we report per-speed success rate alongside the average length of successful rollouts.

As Table~\ref{tab:speed-integration} shows, the three schemes are essentially tied, with overall SRs of $96.8$ / $96.8$ / $96.5$ within $0.3\%$ of each other and comparable rollout lengths at each commanded speed.
This indicates that speed control can be injected into a VLA with little engineering effort, largely independent of the specific mechanism.
Among the three, Text ties for the highest overall SR while requiring no architectural change or pre-defined anchor set, which makes it the simplest and most flexible to deploy.
We therefore adopt the textual prefix as the default speed-integration scheme of TempoVLA in all subsequent experiments.

\subsection{Effect of the Training Speed Range}
\label{sec:sim-range}

We now study how the set of training speeds affects a speed-controllable policy.
Starting from the single-speed baseline, we train three policies on progressively designed speed ranges: a narrow range $\{0.75, 1, 1.25, 1.5\}\times$, a wider range with a larger stride $\{0.5, 1, 1.5, 2\}\times$, and a wide range with a refined stride $\{0.5, 0.75, 1, 1.25, 1.5, 1.75, 2\}\times$.
Each policy is evaluated at every speed it was trained on, and the results are summarized in Table~\ref{tab:speed-range}.

\begin{table}[t]
\centering
\small
\setlength{\tabcolsep}{9.5pt}
\caption{\textbf{Effect of the training speed range on LIBERO.} Each block trains one policy on the indicated speed range and evaluates it at every speed in that range. \emph{Avg.}: success rate averaged over the four suites. \emph{Steps}: average steps of successful rollouts. \emph{Model Ratio}: speed ratio realized by the policy at rollout, measured as $\mathrm{Steps}_{1\times}/\mathrm{Steps}_{s}$. \emph{Data Ratio}: the data-level ratio achieved by VSTA (Section~\ref{sec:sim-feasibility}). Both ratios ideally equal the commanded speed $s$. Red \textcolor{red}{$\uparrow$} marks the $1\times$ gain over the baseline; blue subscripts give the gap (Model Ratio $-$ Data Ratio).}
\vspace{-6pt}
\label{tab:speed-range}
\resizebox{\linewidth}{!}{
\begin{tabular}{l cccc c c cc}
\toprule
Speed & Spatial & Object & Goal & Long & Avg. & Steps & Model Ratio & Data Ratio \\
\midrule
\rowcolor{gray!15}\multicolumn{9}{l}{\textit{Baseline (single-speed)}} \\
$1\times$ & 99.4 & 95.6 & 96.0 & 95.8 & 96.7 & 152 & 1 & 1 \\
\midrule
\rowcolor{gray!15}\multicolumn{9}{l}{\textit{Range} $\{0.75, 1, 1.25, 1.5\}\times$} \\
$0.75\times$ & 99.4 & 95.6 & 97.2 & 93.6 & 96.5 & 197 & 0.77\textsubscript{\textcolor{gapblue}{$+0.01$}} & 0.76 \\
$1.0\times$  & 99.6 & 97.6 & 97.4 & 92.8 & 96.9\textsubscript{\textcolor{red}{$\uparrow$0.2}} & 151 & 1 & 1 \\
$1.25\times$ & 99.4 & 98.0 & 96.6 & 94.0 & 97.0\textsubscript{\textcolor{red}{$\uparrow$0.3}} & 127 & 1.19\textsubscript{\textcolor{gapblue}{$-0.01$}} & 1.20 \\
$1.5\times$  & 99.4 & 97.6 & 96.2 & 94.0 & 96.8\textsubscript{\textcolor{red}{$\uparrow$0.1}} & 111 & 1.36\textsubscript{\textcolor{gapblue}{$-0.07$}} & 1.43 \\
\midrule
\rowcolor{gray!15}\multicolumn{9}{l}{\textit{Range} $\{0.5, 1, 1.5, 2\}\times$} \\
$0.5\times$ & 98.8 & 88.0 & 96.4 & 93.3 & 94.1 & 295 & 0.52\textsubscript{\textcolor{gapblue}{$+0.02$}} & 0.50 \\
$1.0\times$ & 99.2 & 95.2 & 97.8 & 94.8 & 96.8\textsubscript{\textcolor{red}{$\uparrow$0.1}} & 153 & 1 & 1 \\
$1.5\times$ & 99.2 & 98.4 & 96.8 & 94.4 & 97.2\textsubscript{\textcolor{red}{$\uparrow$0.5}} & 111 & 1.38\textsubscript{\textcolor{gapblue}{$-0.05$}} & 1.43 \\
$2.0\times$ & 78.6 & 96.0 & 90.4 & 88.4 & 88.4 & 98  & 1.56\textsubscript{\textcolor{gapblue}{$-0.34$}} & 1.90 \\
\midrule
\rowcolor{gray!15}\multicolumn{9}{l}{\textit{Range} $\{0.5, 0.75, 1, 1.25, 1.5, 1.75, 2\}\times$} \\
$0.5\times$  & 97.6 & 94.4 & 96.0 & 92.1 & 95.0 & 296 & 0.52\textsubscript{\textcolor{gapblue}{$+0.02$}} & 0.50 \\
$0.75\times$ & 98.4 & 95.4 & 97.0 & 94.4 & 96.3 & 201 & 0.76\textsubscript{\textcolor{gapblue}{$\pm 0$}\hphantom{$.01$}} & 0.76 \\
$1.0\times$  & 99.2 & 98.2 & 98.4 & 91.8 & 96.9\textsubscript{\textcolor{red}{$\uparrow$0.2}} & 153 & 1 & 1 \\
$1.25\times$ & 99.0 & 96.0 & 98.8 & 95.6 & 97.4\textsubscript{\textcolor{red}{$\uparrow$0.7}} & 129 & 1.19\textsubscript{\textcolor{gapblue}{$-0.01$}} & 1.20 \\
$1.5\times$  & 98.6 & 98.0 & 96.8 & 95.8 & 97.3\textsubscript{\textcolor{red}{$\uparrow$0.6}} & 112 & 1.37\textsubscript{\textcolor{gapblue}{$-0.06$}} & 1.43 \\
$1.75\times$ & 93.6 & 98.0 & 97.0 & 93.6 & 95.6 & 105 & 1.46\textsubscript{\textcolor{gapblue}{$-0.08$}} & 1.54 \\
$2.0\times$  & 78.6 & 97.0 & 92.4 & 89.6 & 89.4 & 97  & 1.58\textsubscript{\textcolor{gapblue}{$-0.32$}} & 1.90 \\
\bottomrule
\end{tabular}
}
\vspace{-0.15in}
\end{table}

\textbf{Comparison with the baseline.}
Across all three ranges, training with VSTA preserves or improves the $1\times$ success rate over the single-speed baseline ($96.7$).
A per-suite breakdown shows that the gain is concentrated in Object and Goal, both rising by $+2.0$ to $+2.6$ over the baseline.
We attribute this to the speed-conditioned training itself: when the same observation must produce different action magnitudes under different commanded speeds, the policy can no longer memorize a single observation-to-magnitude mapping, and is forced to extract finer object- and goal-aware features that also transfer to the $1\times$ regime.

\textbf{Peak performance shifts away from $1\times$.}
More strikingly, the peak success rate of every speed-conditioned policy occurs not at $1\times$ but at $1.25\times$ or $1.5\times$: $97.0$, $97.2$, and $97.4$ for the narrow, four-speed, and seven-speed ranges respectively, each exceeding its $1\times$ counterpart.
We attribute this to natural pacing slack in teleoperation data: even on the clean LIBERO benchmark, demonstrations contain rhythm padding and ambiguous transition frames that VSTA's merge operation compresses out at moderate speedups.
Trained under this compression, the policy executes more decisively at $1.25\times$ and $1.5\times$, which reduces the ambiguity-induced stalls that occasionally appear at the original $1\times$ rate.
A practical implication is that the default deployment speed of TempoVLA is best set slightly above $1\times$ rather than at the demonstration rate.

\textbf{Effect of the speed range.}
Comparing the ranges reveals two consistent trends.
First, a finer speed granularity helps: over the shared speeds $\{0.5, 1, 1.5, 2\}\times$, refining the stride from $0.5$ to $0.25$ raises the success rate at every speed (e.g., $94.1\!\to\!95.0$ at $0.5\times$ and $88.4\!\to\!89.4$ at $2\times$).
Second, including the extreme $0.5\times$ and $2\times$ broadens the augmentation enough to lift the moderate speeds, where the seven-speed range matches or exceeds the narrow $\{0.75\text{--}1.5\}\times$ range at $1$, $1.25$, and $1.5\times$.
The refined seven-speed range thus offers the best overall trade-off between coverage and granularity.

\textbf{Realized versus data speed ratio.}
Finally, we compare the Model Ratio (the speed actually realized at rollout) with the Data Ratio (the speed achievable on the augmented data).
The model broadly hits the target but under-shoots at high speedups (e.g., $1.56\times$ realized at the $2\times$ command versus a $1.90\times$ data ratio).
The gap arises from two factors: corrective steps after imperfect first attempts inflate the rollout length, and the low-level controller cannot accurately track the large action magnitudes.


\section{Real-world Experiments}
\label{sec:real-exp}

\begin{figure}[t]
  \centering
  \includegraphics[width=1.01\linewidth]{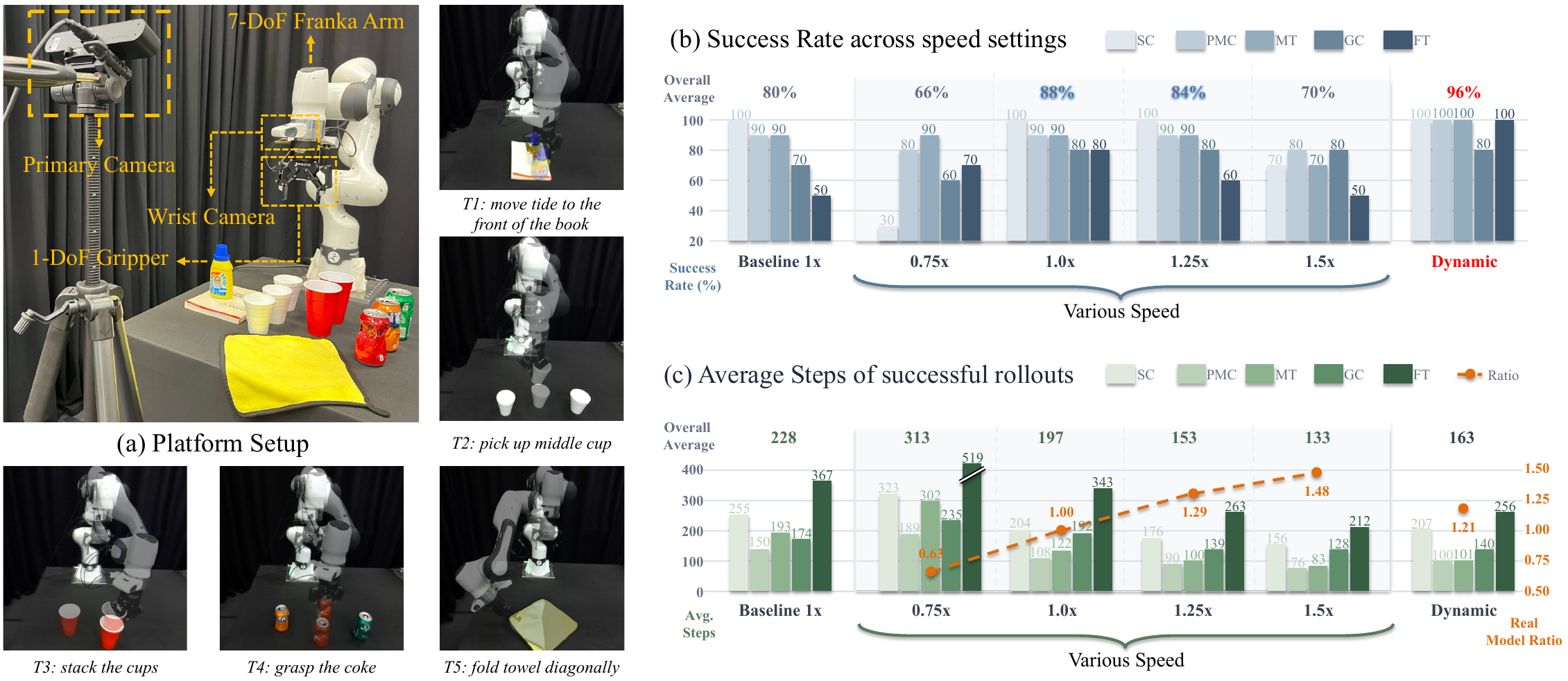}
  \vspace{-12pt}
  \caption{\textbf{Real-world Setup and Results.} \textit{(a)} Four pick-and-place behaviors and one deformable-object task with Franka. \textit{(b)} TempoVLA improves $1\times$ success rate from $80\%$ to $88\%$ over single-speed baseline, and the GPT-4o-scheduled variant reaches the highest overall success rate of $96\%$. \textit{(c)} The realized Model Ratio closely tracks the commanded ratio for $s$ = ($1.29$, $1.48$).}
  \label{fig:real-tasks}
\end{figure}

We deploy TempoVLA on a $7$-DoF Franka arm with a $1$-DoF parallel gripper, observed by a primary camera and a wrist-mounted camera (Figure~\ref{fig:real-tasks}).
We evaluate on five tasks covering four pick-and-place behaviors and one deformable-object task.
For each task we collect $50$ teleoperated trajectories for training and run $10$ rollouts per commanded speed for evaluation.
At inference, the policy re-queries after executing the first $10$ steps of each predicted action chunk.
Each action is $8$-dimensional, consisting of a $7$-dim joint velocity and a $1$-dim gripper value, both lying in a linearly composable space so that VSTA applies directly.
We train one TempoVLA policy on the speed set $\{0.75, 1, 1.25, 1.5\}\times$ alongside a single-speed baseline at $1\times$ for comparison.

\subsection{Results}
\label{sec:real-results}

\noindent\textbf{VSTA boosts the $1\times$ success rate, mirroring the simulation finding.}
On the Franka platform, VSTA boosts the default $1\times$ success rate from $80.0$ (single-speed baseline) to $88.0$, an $8$-point gain that mirrors the implicit-augmentation effect observed in simulation.
The $1.25\times$ speed also outperforms the baseline ($84.0$ versus $80.0$), showing that TempoVLA delivers consistent gains at and above the demonstration speed.

\noindent\textbf{Realized speedup closely matches the commanded ratio.}
The Model Ratio realized at rollout tracks the commanded speed across the trained range, with $0.63\times$, $1.29\times$, and $1.48\times$ realized at commanded $0.75\times$, $1.25\times$, and $1.5\times$ respectively.
This confirms that TempoVLA's speed conditioning translates faithfully into execution-speed control on real hardware, not only at the policy-prediction level but also through the unchanged low-level controller.

\subsection{Dynamic Speed Control with a VLM Scheduler}
\label{sec:real-dynamic}

We further test whether TempoVLA, paired with a high-level VLM, can schedule its own speed at runtime (Figure~\ref{fig:framework}\,(c)).
We adopt GPT-4o~\citep{hurst2024gpt4o} as the scheduler, querying it once every two action chunks to dispatch the speed for the next segment.
TempoVLA with dynamic scheduling reaches $96\%$ average success rate, $8$ points above the best fixed-speed configuration ($88\%$ at $1\times$), while still completing tasks at an average realized speedup of $1.21\times$ over the $1\times$ baseline.

In our prompt (full text in Appendix~\ref{sec:appendix-gpt-prompt}) we explicitly encourage GPT-4o to favor aggressive speedups during low-risk phases.
Yet the actual schedule remains conservative, with the vast majority of decisions falling on the $1\times$ or $1.25\times$ tier and $1.5\times$ rarely dispatched.
Despite this conservatism, GPT-4o reads the execution state of the real robot with high reliability, correctly anticipating free-space transit, fine alignment, and contact phases.
The resulting schedule realizes the phase-aware variable-speed behavior we expect, just biased toward the safer end of the speed range.

\section{Conclusion}
\label{sec:conclusion}

Existing Vision-Language-Action models inherit a single fixed execution speed from training data.
We propose TempoVLA, a single speed-controllable VLA framework that pairs a data-side Variable-Speed Trajectory Augmentation with a lightweight model-side conditioning mechanism, both lightweight and applicable to existing VLAs.
Experiments in simulation and real world show that TempoVLA delivers flexible bidirectional speed control ability, improves default $1\times$ performance, and achieves dynamic speed control with an external VLM.

\textbf{Limitation and Future Work.}
At the high end of the speed range, the realized speedup gradually saturates because the policy's per-step targets begin to exceed the fixed low-level controller's tracking bandwidth (Appendix~\ref{sec:sim-extreme}), and co-tuning the controller alongside TempoVLA is a natural extension (Appendix~\ref{sec:future}).


\clearpage


\bibliography{main}  

\newpage
\appendix
\section{Hyperparameters}
\label{sec:appendix-train-hp}

\begin{table}[h]
\centering
\begin{minipage}[t]{0.48\textwidth}
\centering
\caption{Training hyperparameters of $\pi_{0.5}$ on LIBERO.}
\label{tab:hp-libero}
\begin{tabular}{lc}
\toprule
\textbf{Hyperparameter} & \textbf{Value} \\
\midrule
GPUs              & 32 $\times$ H20 \\
Total Batch Size  & 512 \\
Optimizer         & AdamW \\
Scheduler         & Cosine Decay \\
Learning Rate     & 5e-5 \\
Iterations        & 30k \\
Warmup Step       & 1k \\
Minimum LR        & 5e-7 \\
Max Gradient Norm & 1.0 \\
\bottomrule
\end{tabular}
\end{minipage}
\hfill
\begin{minipage}[t]{0.48\textwidth}
\centering
\caption{Training hyperparameters of $\pi_{0.5}$ on real-world tasks.}
\label{tab:hp-real}
\begin{tabular}{lc}
\toprule
\textbf{Hyperparameter} & \textbf{Value} \\
\midrule
GPUs              & 16 $\times$ H20 \\
Total Batch Size  & 128 \\
Optimizer         & AdamW \\
Scheduler         & Cosine Decay \\
Learning Rate     & 5e-5 \\
Iterations        & 10k \\
Warmup Step       & 1k \\
Minimum LR        & 5e-7 \\
Max Gradient Norm & 1.0 \\
\bottomrule
\end{tabular}
\end{minipage}
\end{table}

\noindent\textbf{Training hyperparameters.}
Please refer to Table~\ref{tab:hp-libero} and Table~\ref{tab:hp-real}.
When evaluating at slow commanded speeds, we proportionally raise the maximum rollout step budget so that the policy is given enough time to complete the task at the reduced per-step magnitude.

\noindent\textbf{VSTA segmentation hyperparameters.}
For VSTA segmentation on the real-robot data, we mark a frame as a direction change when consecutive action directions diverge by more than $60^\circ$, while on the LIBERO, the threshold is $90^\circ$.
For both environments, the gripper event are labeled when the absolute gripper state changes by more than $0.5$.
This partitions each demonstration into $5.2/6.0$ segments on average with a mean length of $45/41$ steps in real-world tasks/LIBERO.

\section{Pseudocode of VSTA}
\label{sec:appendix-vsta-alg}

Algorithm~\ref{alg:vsta} gives the full pseudocode of Variable-Speed Trajectory Augmentation, applied online to a single demonstration during training.

\begin{algorithm}[h]
\caption{Online Variable-Speed Trajectory Augmentation (one demonstration)}
\label{alg:vsta}
\begin{algorithmic}[1]
\Require demonstration $\tau=\{(o_t, a_t)\}_{t=1}^{T}$, target speed $s$
\State write $s = q/p$ with coprime integers $q, p$ \Comment{$q$ source $\rightarrow$ $p$ output frames}
\State $\{S_k\} \gets \textsc{Segment}(\tau)$ \Comment{motion mode + direction split; gripper events as anchors}
\State $\mathcal{T} \gets \varnothing$ \Comment{re-timed (action, validity) stream, in output order}
\For{each segment $S_k$}
    \State sample chunk-start offset $r \sim \mathcal{U}\{0,\dots,q-1\}$ \Comment{drawn online, per segment}
    \State append frames $[0, r)$ to $\mathcal{T}$, all marked valid \Comment{leading passthrough}
    \For{each non-overlapping chunk of $q$ frames starting at $r$}
        \State $\Delta \gets \textstyle\sum_{i} a_i$ over the $q$ frames \Comment{accumulate motion}
        \State re-split $\Delta$ into $p$ steps by interpolating the cumulative motion
        \State append the $p$ steps to $\mathcal{T}$, marking only the chunk-start observation valid
    \EndFor
    \State append the trailing ${<}\,q$ leftover frames to $\mathcal{T}$, all marked valid \Comment{trailing passthrough}
\EndFor
\State \Return re-timed trajectory $\mathcal{T}$ with its validity mask
\end{algorithmic}
\end{algorithm}

\section{More Ablation Study}

\subsection{The Effect of Soft Prompt Length}

As Table~\ref{tab:soft-prompt-length} shows, the average success rate is essentially flat across $P \in \{4, 8, 16\}$, with the three averages all within $0.3$ points of each other.
Performance only starts to slip at $P=32$ ($96.3$ on average), suggesting that a few anchor tokens per speed are sufficient and that over-long prompts mildly hurt optimization.
We therefore adopt $P=8$ as the default in our experiments.

\begin{table}[h]
\centering
\footnotesize
\setlength{\tabcolsep}{9pt}
\caption{\textbf{Effect of soft prompt length $P$ on LIBERO.} All variants use the speed set $\{0.75, 1, 1.25, 1.5\}\times$. \emph{SR}: average success rate (\%); \emph{Steps}: average rollout length on successes.}
\label{tab:soft-prompt-length}
\begin{tabular}{l cc cc cc cc}
\toprule
\multirow{2}{*}{Speed} & \multicolumn{2}{c}{$P=4$} & \multicolumn{2}{c}{$P=8$} & \multicolumn{2}{c}{$P=16$} & \multicolumn{2}{c}{$P=32$} \\
\cmidrule(lr){2-3} \cmidrule(lr){4-5} \cmidrule(lr){6-7} \cmidrule(lr){8-9}
 & SR\,$\uparrow$ & Steps & SR\,$\uparrow$ & Steps & SR\,$\uparrow$ & Steps & SR\,$\uparrow$ & Steps \\
\midrule
$0.75\times$ & 96.4 & 199 & 96.4 & 200 & 96.9 & 199 & 95.8 & 199 \\
$1.0\times$  & 96.8 & 153 & 95.8 & 153 & 96.9 & 152 & 96.7 & 152 \\
$1.25\times$ & 96.7 & 128 & 96.8 & 129 & 96.5 & 127 & 97.1 & 128 \\
$1.5\times$  & 96.7 & 112 & 96.9 & 111 & 97.0 & 111 & 95.7 & 111 \\
\midrule
Avg. & 96.6 & 147 & 96.5 & 148 & 96.8 & 147 & 96.3 & 147 \\
\bottomrule
\end{tabular}
\vspace{-0.15in}
\end{table}

\section{Stress Test at Extreme Speeds}
\label{sec:sim-extreme}

To probe the limits of TempoVLA, we train a single policy on the wide, fine-grained range $\{0.25, 0.5, 0.75, 1, 1.5, 2, 2.5, 3, 4\}\times$ and evaluate it at every training speed.
Table~\ref{tab:speed-extreme} reports the per-speed success rate together with the realized Model Ratio, the Data Ratio, and the controller's tracking gap.

\begin{table}[h]
\centering
\small
\setlength{\tabcolsep}{12pt}
\caption{\textbf{Performance under an extreme speed range (LIBERO).} A single TempoVLA policy trained on the wide range $\{0.25, 0.5, 0.75, 1, 1.5, 2, 2.5, 3, 4\}\times$ and evaluated across speeds. \emph{Controller Gap}: tracking error between the per-step EEF motion requested of the controller and the motion actually realized in one simulation step. Blue subscripts give Model Ratio $-$ Data Ratio.}
\label{tab:speed-extreme}
\begin{tabular}{l c c c c cc}
\toprule
\multirow{2}{*}{Speed} & \multirow{2}{*}{Avg.} & \multirow{2}{*}{Steps} & \multirow{2}{*}{Model Ratio} & \multirow{2}{*}{Data Ratio} & \multicolumn{2}{c}{Controller Gap} \\
\cmidrule(lr){6-7}
 & & & & & Pos.\,(m) & Rot.\,(rad) \\
\midrule
$0.25\times$ & 75.8 & 536 & 0.28\textsubscript{\textcolor{gapblue}{$+0.03$}} & 0.25 & 0.009 & 0.015 \\
$0.5\times$  & 92.8 & 288 & 0.53\textsubscript{\textcolor{gapblue}{$+0.03$}} & 0.50 & 0.019 & 0.031 \\
$1\times$    & 96.6 & 152 & 1.00 & 1.00 & 0.038 & 0.069 \\
$1.5\times$  & 96.6 & 112 & 1.36\textsubscript{\textcolor{gapblue}{$-0.07$}} & 1.43 & 0.057 & 0.100 \\
$2\times$    & 88.3 & 95  & 1.60\textsubscript{\textcolor{gapblue}{$-0.30$}} & 1.90 & 0.075 & 0.129 \\
$2.5\times$  & 72.8 & 96  & 1.58\textsubscript{\textcolor{gapblue}{$-0.51$}} & 2.09 & 0.095 & 0.151 \\
$3\times$    & 56.0 & 99  & 1.54\textsubscript{\textcolor{gapblue}{$-1.05$}} & 2.59 & 0.112 & 0.189 \\
$4\times$    & 34.3 & 93  & 1.63\textsubscript{\textcolor{gapblue}{$-1.43$}} & 3.06 & 0.146 & 0.243 \\
\bottomrule
\end{tabular}
\end{table}

\noindent\textbf{Speed control degrades gracefully within $0.5\times$ to $1.5\times$ and breaks beyond it.}
Inside this regime, SR stays above $92$ and the realized Model Ratio closely tracks the target.
Outside, performance degrades on both ends.
At $0.25\times$, SR drops to $75.8$ because the per-step magnitudes shrink to nearly zero and the policy becomes sensitive to ambiguous observations.
At the high end, SR collapses from $96.6$ at $1.5\times$ to $34.3$ at $4\times$, and the realized Model Ratio saturates around $1.6$, far below the commanded speed.

\noindent\textbf{The acceleration bottleneck is the controller, not TempoVLA.}
The \emph{Controller Gap} columns measure the discrepancy between the per-step end-effector target sent to the controller and the motion actually realized after one simulation step.
This gap grows steeply with speed, from $0.038$\,m\,/\,$0.069$\,rad at $1\times$ to $0.146$\,m\,/\,$0.243$\,rad at $4\times$, indicating that the per-step target becomes too large for the operational-space controller and the robot dynamics to realize within one control interval.
Action clipping stays near zero throughout, ruling out the controller's input range as the limiting factor.
The Model Ratio saturating around $1.6$ for $s \ge 2\times$ is a direct consequence: no matter what TempoVLA predicts, the robot cannot move faster than the controller can track.

\noindent\textbf{Summary.}
TempoVLA itself extends gracefully across $0.5\times$ to $1.5\times$, while headroom beyond this regime is governed by the low-level controller rather than the policy.
Reaching higher real speeds therefore requires joint tuning of the controller alongside TempoVLA, consistent with the orthogonality view in Section~\ref{sec:relationship}.

\section{Qualitative Comparison}

\subsection{Failure Mode Analysis}
\label{sec:appendix-failure}

\begin{figure}[t]
  \centering
  \includegraphics[width=0.99\linewidth]{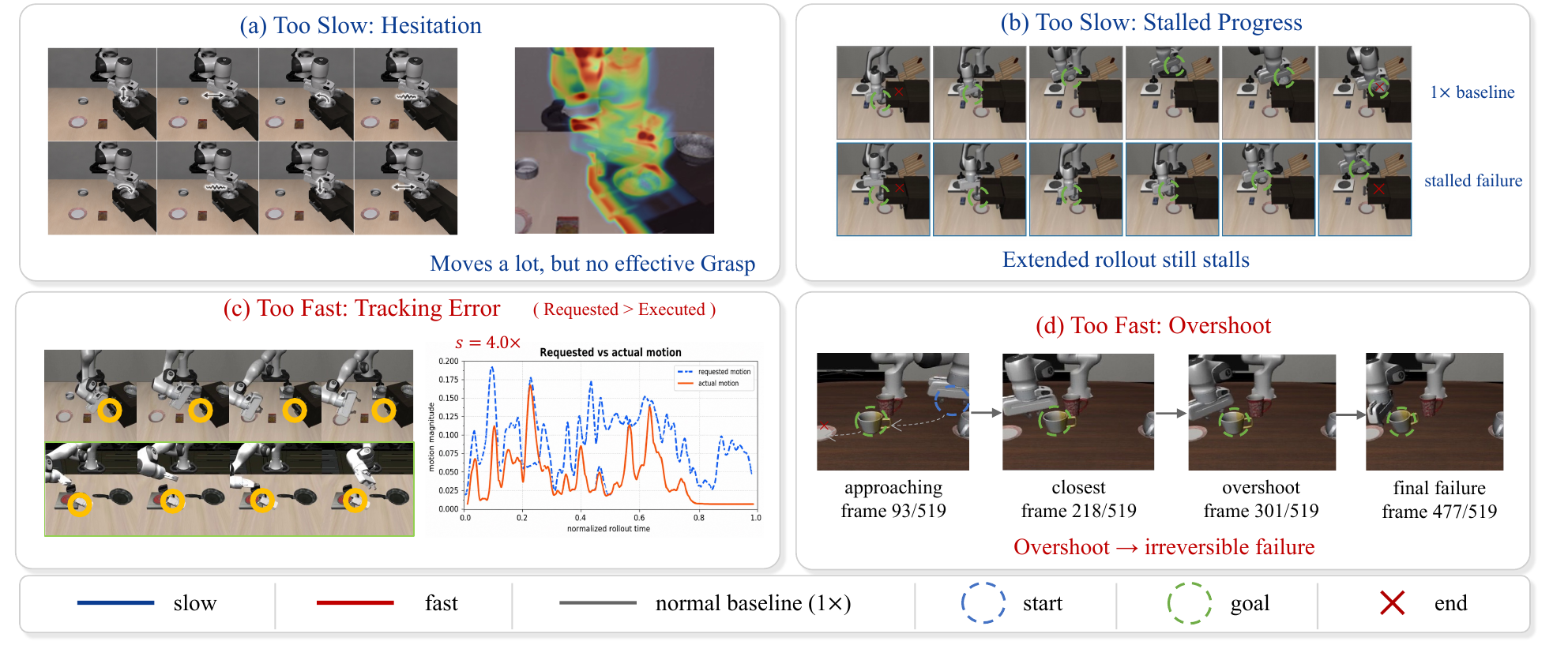}
  \caption{\textbf{Failure Mode Analysis.}}
  \label{fig:failure_mode}
  \vspace{-0.1in}
\end{figure}

\noindent\textbf{Qualitative failure mode analysis.}
Figure~\ref{fig:failure_mode} provides qualitative examples for the degradation patterns observed in the extreme-speed stress test.
This analysis is not intended to suggest that TempoVLA is unreliable within its practical operating range.
As shown in Table~\ref{tab:speed-extreme}, the policy remains robust for moderate commands between $0.5\times$ and $1.5\times$, while performance degrades primarily near the two ends of the evaluated speed spectrum.
The examples below therefore characterize the boundary cases of speed-conditioned execution.

\noindent\textbf{Slow-speed failures.}
At very low speeds, the dominant failure mode is insufficient task progress.
In Figure~\ref{fig:failure_mode}(a), the end effector produces repeated local motions around the target object, but these motions do not accumulate into an effective grasp.
This is distinct from a static failure: the policy remains active, yet the reduced per-step displacement is too small to reliably drive the system across key manipulation transitions, such as approach-to-contact and contact-to-grasp.
We refer to this behavior as \emph{hesitation}.
It is consistent with the quantitative result at $0.25\times$, where action magnitudes approach zero and the policy becomes more sensitive to visually ambiguous states.

Figure~\ref{fig:failure_mode}(b) shows a related stalled-progress failure.
Even with an extended rollout horizon, the robot remains in a similar local behavior pattern and does not complete the task.
This indicates that the failure is not merely due to an insufficient number of control steps.
Rather, each step contributes too little effective progress, allowing the policy to remain trapped near a phase boundary instead of transitioning to the next manipulation stage.
Thus, slow execution is beneficial only when the reduced action magnitude still preserves enough progress to complete the required phase transition.

\noindent\textbf{Fast-speed failures.}
At high speeds, failures arise from a different mechanism.
Figure~\ref{fig:failure_mode}(c) shows a mismatch between requested and realized motion: the policy issues larger per-step targets, but the low-level controller cannot faithfully execute them within one control interval.
This observation matches the controller tracking-gap measurements in Table~\ref{tab:speed-extreme}, where the realized Model Ratio saturates around $1.6\times$ even as the commanded speed continues to climb.
This suggests the high-speed limit is primarily imposed by execution-side tracking rather than by errors introduced by VSTA.

Figure~\ref{fig:failure_mode}(d) illustrates a downstream consequence of this tracking mismatch.
The end effector approaches the target region too aggressively, passes the valid interaction window, and fails before the policy can correct its motion.
We refer to this failure mode as \emph{overshoot}.
Such failures are particularly damaging in contact-rich stages, where successful interaction often depends on a narrow spatial and temporal tolerance.
Once the gripper moves past the object or perturbs it into an out-of-distribution state, the remaining rollout can become unrecoverable.

\noindent\textbf{Implication.}
These qualitative examples clarify the usable speed envelope of TempoVLA.
Slow commands can lead to hesitation or stalled progress because each action contributes too little effective task progress.
Fast commands can lead to tracking error or overshoot because the requested per-step motion exceeds the controller's tracking capability.
TempoVLA should therefore not be used by assigning an extreme fixed speed to the entire rollout.
Instead, speed should be selected according to the current manipulation phase: faster during low-risk free-space motion and slower near contact-rich phases that require precise interaction.
This observation is consistent with our dynamic speed scheduling results, where phase-aware speed selection outperforms fixed-speed execution.

\subsection{Demonstration}

\begin{figure}[p]
  \centering
  \includegraphics[width=\linewidth,height=0.46\textheight,keepaspectratio]{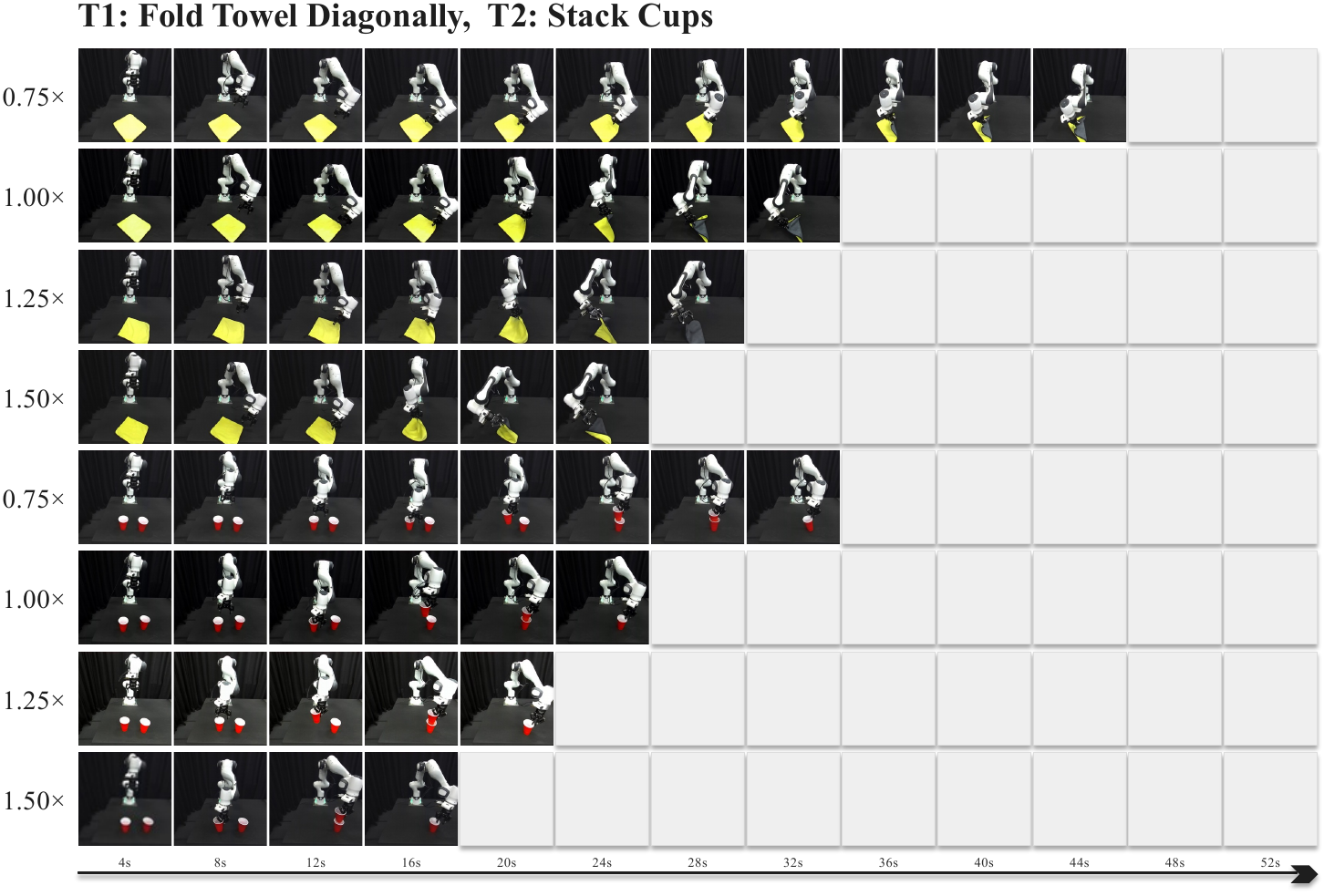}\\[1em]
  \includegraphics[width=\linewidth,height=0.46\textheight,keepaspectratio]{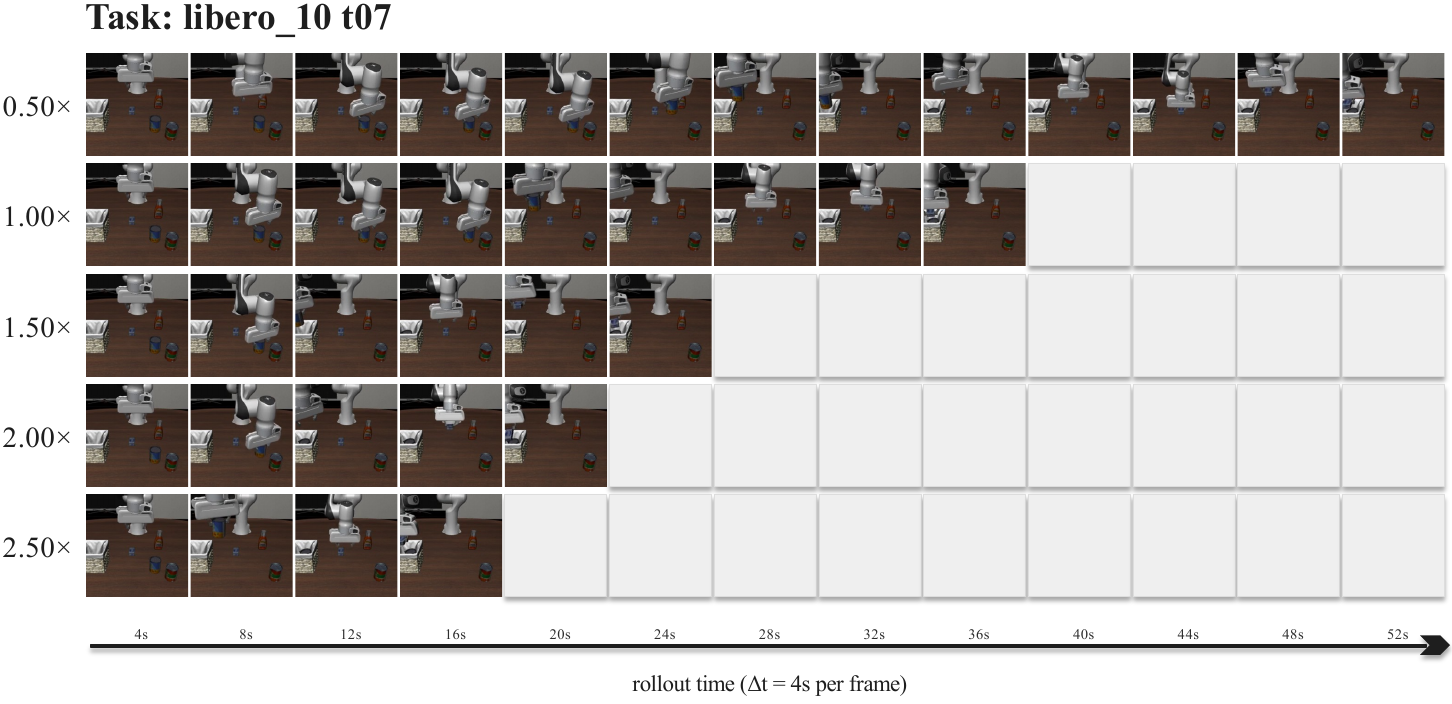}
  \caption{\textbf{Demonstration rollouts of TempoVLA at various speed (1/2).}}
  \label{fig:demo-part1}
\end{figure}

\begin{figure}[p]
  \centering
  \includegraphics[width=\linewidth,height=0.30\textheight,keepaspectratio]{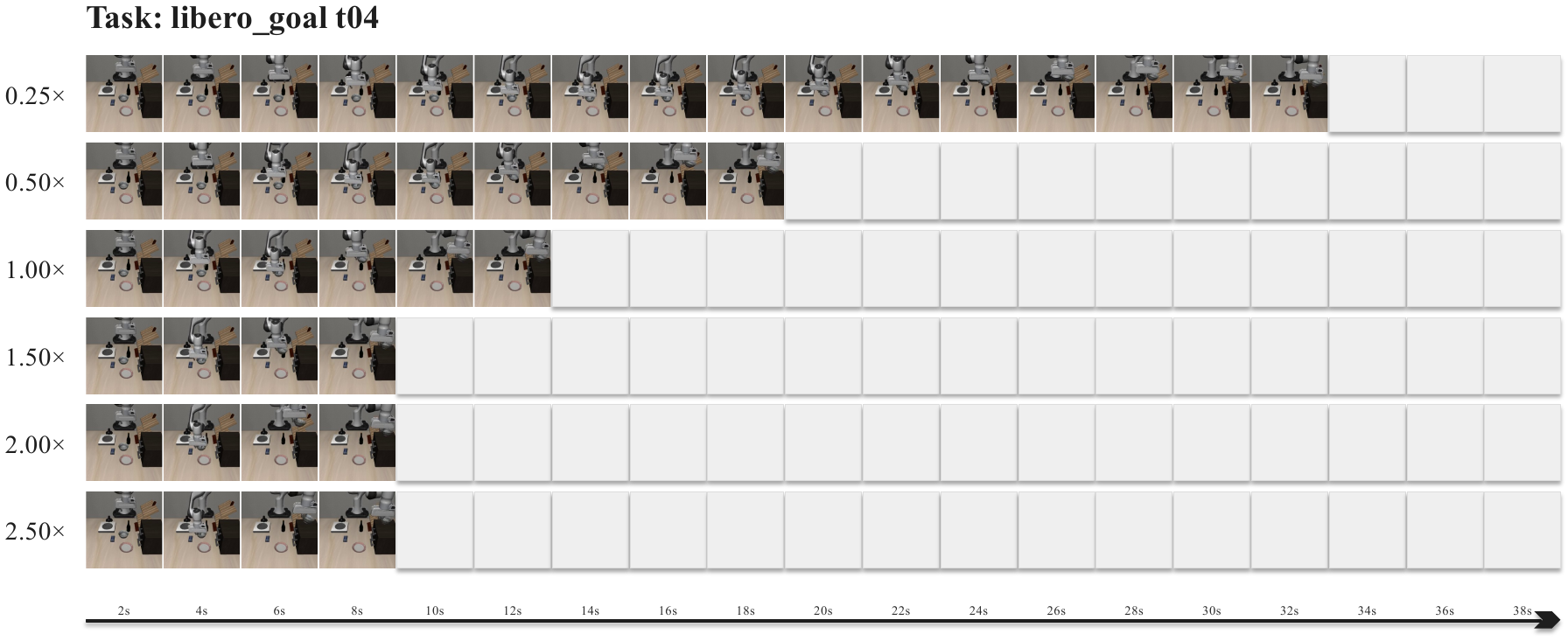}\\[1em]
  \includegraphics[width=\linewidth,height=0.30\textheight,keepaspectratio]{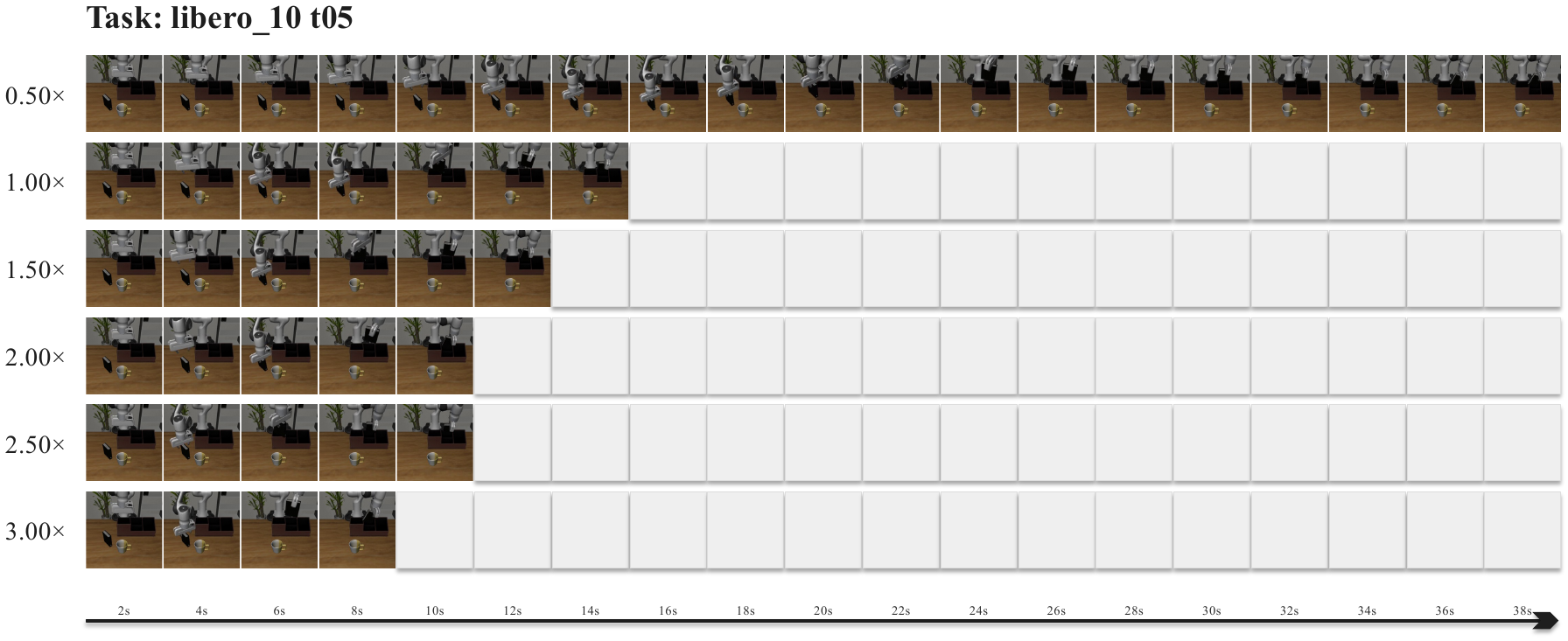}\\[1em]
  \includegraphics[width=\linewidth,height=0.30\textheight,keepaspectratio]{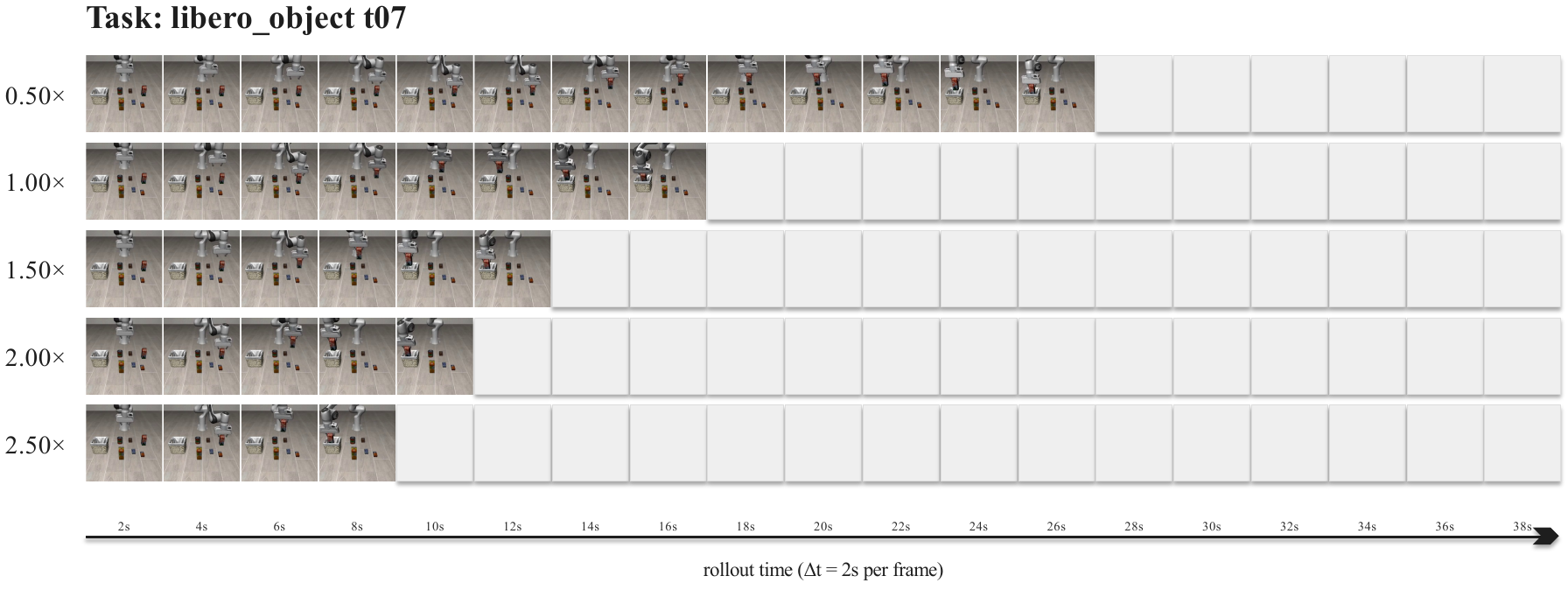}
  \caption{\textbf{Demonstration rollouts of TempoVLA at various speed (2/2).}}
  \label{fig:demo-part2}
\end{figure}

In Figures~\ref{fig:demo-part1} and~\ref{fig:demo-part2}, we present demonstration rollouts at different speeds and tasks for illustration.

\section{Prompt to GPT4o for Dynamic Speed Control}
\label{sec:appendix-gpt-prompt}
\vspace{-0.1in}

\begin{figure}[p]
  \centering
  \includegraphics[width=\linewidth,height=0.92\textheight,keepaspectratio]{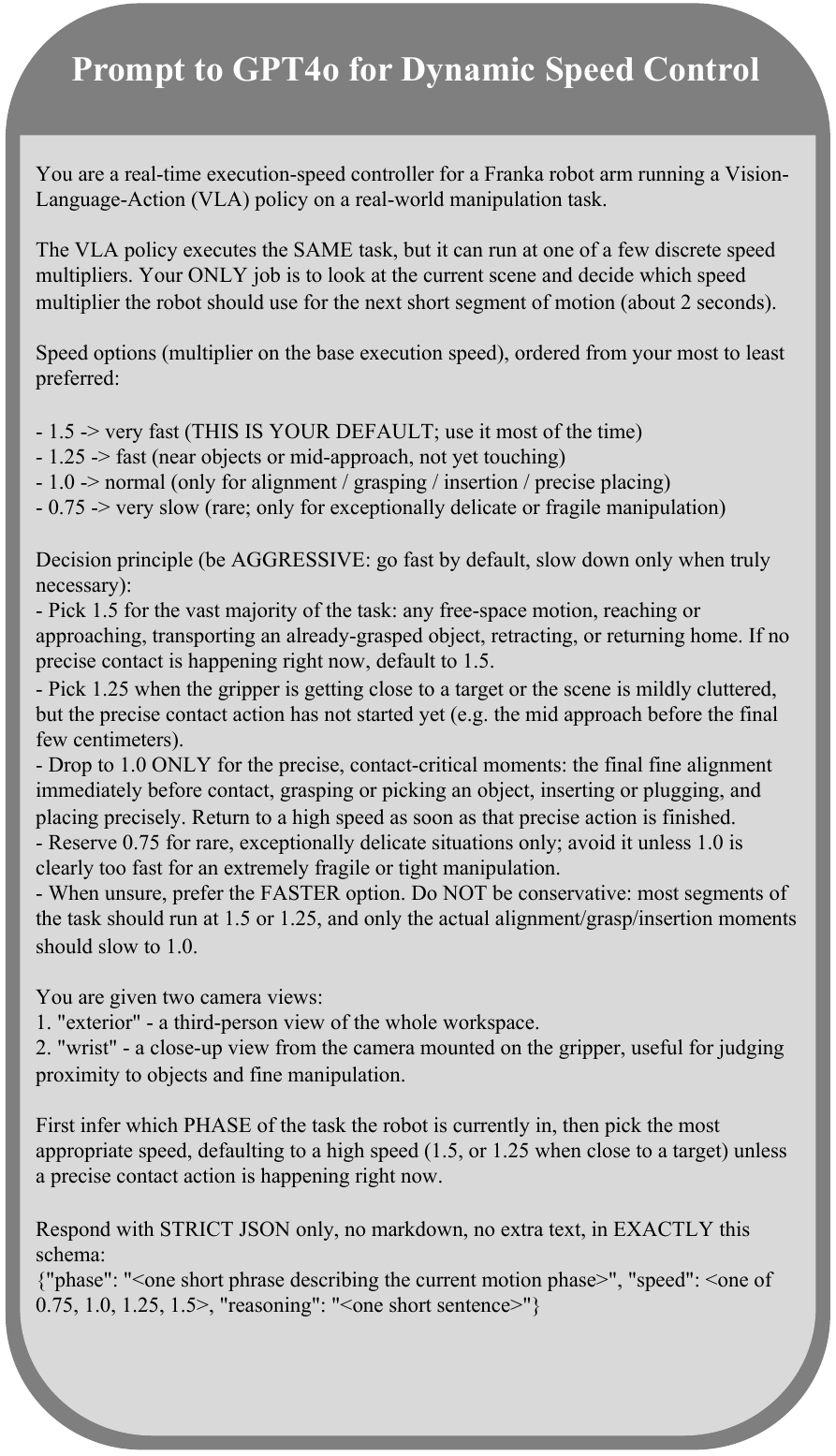}
  \caption{\textbf{Prompt sent to GPT-4o for dynamic speed control.}}
  \label{fig:gpt-prompt}
\end{figure}

Figure~\ref{fig:gpt-prompt} reproduces the prompt we send to GPT-4o for dispatching the per-segment speed during dynamic speed control.

\section{Discussion}
\vspace{-0.1in}

\subsection{Relationship with Controller-focused Works}
\label{sec:relationship}

A natural alternative to changing the execution speed of a robot is to act on the low-level controller, for example by scaling target velocities or stretching step periods after the policy has produced its outputs.
We view such controller-side methods as orthogonal to TempoVLA rather than competing with it: controllers sit downstream of the policy and can only rescale or retime what the policy has already predicted, while variable-speed training changes the content of those predictions at the policy level itself.
The two approaches therefore intervene at different layers of the control stack and can be composed directly, so a TempoVLA policy can still be paired with any controller-side modulation when finer execution-side tuning is desired.

\subsection{The difference between effect of VSTA on EEF and Joint Action Space}
\label{sec:vsta-action-space}

For manipulation tasks, success is defined by the relative pose between the end effector and the manipulated objects, so the end-effector trajectory is the quantity we ultimately care about when evaluating how faithfully a re-timed demonstration preserves the original task semantics.
Both EEF and joint commands satisfy the linear-composability requirement of Section~\ref{sec:method-vsta}, so VSTA's accumulate-then-split operation is mathematically well-defined in either case.
The relevant question is therefore not whether VSTA is valid on joints, but how large the resulting end-effector deviation becomes after re-timing, and three structural reasons make this deviation larger in joint space than in EEF space.

\noindent\textbf{Geometric amplification at single joints.}
A small re-timing error on a proximal joint is geometrically magnified at the end effector in proportion to its lever arm, a structural penalty that joint-space re-timing cannot avoid.

\noindent\textbf{Non-linear coupling across the kinematic chain.}
Forward kinematics maps joint angles to end-effector pose through a non-linear function, so linearly interpolated joint increments translate into end-effector motions that no longer interpolate linearly, while re-timing directly on EEF translations and axis-angle increments keeps the linear interpolation aligned with the quantity that defines task success.

\noindent\textbf{Controller realizability under speed change.}
For operational-space controllers commonly used on modern arms, joint-space re-timing commits the policy to a specific joint trajectory, whereas EEF-space re-timing lets the controller use its inverse-kinematics resolution to find a joint trajectory dynamically feasible at the new speed, helping absorb effects such as inertia, friction, and torque saturation.

Taken together, these reasons explain why we prefer EEF actions for VSTA when both representations are available, although VSTA can still be applied directly on platforms that expose only joint commands without harming task-relevant performance in our setup.

\section{Future Directions}
\label{sec:future}

\noindent\textbf{Extending VSTA to non-composable action spaces.}
VSTA's current implementation assumes the action space is closed under linear composition, which excludes representations such as unit quaternions, rotation matrices, and Euler angles.
A one-time mapping to a tangent-space representation or an on-manifold interpolation scheme such as SLERP~\citep{shoemake1985animating} extends VSTA to these representations without changing its core algorithm, broadening the plug-and-play scope of TempoVLA across more platforms.

\noindent\textbf{Co-tuning TempoVLA with the low-level controller.}
At the high end of the speed range, the realized speedup is bounded by the fixed low-level controller rather than by TempoVLA itself, since we deliberately leave the controller unchanged throughout this work to isolate the contribution of policy-level speed control.
Pairing TempoVLA with controller-side adjustments such as a higher control frequency, a wider admissible action range, or finer sub-step action decomposition would push this ceiling further and let the policy's full speed-control envelope translate into even larger executed speedups, consistent with the orthogonality view in Section~\ref{sec:relationship}.

\noindent\textbf{Reducing the VLM scheduling latency.}
In our current dynamic speed control implementation, the GPT-4o scheduler is invoked synchronously between action chunks, which adds wall-clock overhead to the rollout.
This cost can be hidden by feeding the scheduler a longer observation history and running it asynchronously in parallel with TempoVLA, so that scheduling decisions arrive in time for the next chunk without blocking inference.
A thorough exploration of this engineering optimization is left to future work.

\noindent\textbf{Default speed regularization.}
TempoVLA currently treats the original demonstration speed as $1\times$, which implicitly assumes that the per-action granularity within a dataset is uniform.
In practice, however, action granularity varies considerably across demonstrations and even across segments within the same demonstration, since human operators rarely move at a strictly constant pace.
A cleaner formulation would first apply a VSTA-style normalization to flatten this within-dataset speed variability before defining the $1\times$ reference, so that the speed scalar $s$ would condition the policy on a deviation from a well-calibrated mean rather than from an inconsistent demonstrator pace.
We expect this to sharpen the correspondence between the commanded speed and the realized execution speed, and leave a principled implementation of this normalization step to future work.

\end{document}